\documentclass[10pt,twocolumn,letterpaper]{article}

\usepackage{cvpr} 

\usepackage[accsupp]{axessibility} 

\usepackage{url}
\usepackage{float}
\usepackage{makecell}
\usepackage{colortbl}
\usepackage{xcolor}
\usepackage{booktabs}
\usepackage{graphicx}
\definecolor{cvprblue}{rgb}{0.21,0.49,0.74}
\usepackage[colorlinks,linkcolor=cvprblue,citecolor=cvprblue,urlcolor=cvprblue]{hyperref}

\def\paperID{32}
\def\confName{CVPR}
\def\confYear{2026}

\newcommand{\method}{ZACH-ViT}

\title{Extending \method{} to Robust Medical Imaging: Corruption and Adversarial Stress Testing in Low-Data Regimes}

\author{
Athanasios Angelakis\\
BioML Lab, RI CODE, UniBw, Munich, Germany\\
AUMC, Amsterdam, Netherlands\\
\texttt{athanasios.angelakis@unibw.de}
\and
Marta Gomez-Barrero\\
BioML Lab, RI CODE, UniBw, Munich, Germany\\
\texttt{marta.gomez-barrero@unibw.de}
}

\date{}

\begin{document}
\maketitle
\thispagestyle{empty}
\pagestyle{empty}

\begin{abstract}
The recently introduced \method{} (Zero-token Adaptive Compact Hierarchical Vision Transformer) formalized a compact permutation-invariant Vision Transformer for medical imaging and argued that architectural alignment with spatial structure can matter more than universal benchmark dominance. Its design was motivated by the observation that positional embeddings and a dedicated class token encode fixed spatial assumptions that may be suboptimal when spatial organization is weakly informative, locally distributed, or variable across biomedical images. The foundational study established a regime-dependent clean performance profile across MedMNIST, but did not examine robustness in detail. In this work, we present the first robustness-focused extension of \method{} by evaluating its behavior under common image corruptions and adversarial perturbations in the same low-data setting. We compare \method{} with three scratch-trained compact baselines, ABMIL, Minimal-ViT, and TransMIL, on seven MedMNIST datasets using 50 samples per class, fixed hyperparameters, and five random seeds. Across the benchmark, \method{} achieves the best overall mean rank on clean data (1.57) and under common corruptions (1.57), indicating a favorable balance between baseline predictive performance and robustness to realistic image degradation. Under adversarial stress, all models deteriorate substantially; nevertheless, \method{} remains competitive, ranking first under FGSM (2.00) and second under PGD (2.29), where ABMIL performs best overall. These results extend the original \method{} narrative: the advantages of compact permutation-invariant transformers are not limited to clean evaluation, but can persist under realistic perturbation stress in low-data medical imaging, while adversarial robustness remains an open challenge for all evaluated models.

Accepted at CVPR 2026 Workshop (PHAROS-AIF-MIH)
\end{abstract}

\section{Introduction}
Vision Transformers are now widely used in computer vision, but their standard formulation still assumes that spatial order should be encoded explicitly through positional embeddings and token-based aggregation. The recently introduced \method{} \cite{angelakis2026zachvit} questioned whether this assumption is always appropriate for medical imaging, where diagnostically relevant structure may be weakly ordered, locally distributed, or variable across acquisitions. In that setting, the foundational \method{} study argued that architectural inductive biases should align with the spatial characteristics of the target data rather than assume uniform spatial relevance, and showed a regime-dependent clean performance profile under strict few-shot training.

That first paper primarily established the architectural motivation and the clean performance profile of the model. A related question remained open: \emph{does the same inductive-bias choice remain advantageous when robustness is evaluated directly?} This question is practically relevant because robustness has become a central concern in healthcare machine learning, where performance may shift under acquisition variability, data-quality degradation, and other departures from development conditions \cite{balendran2025robustness}. Realistic failures can arise from blur, compression artifacts, brightness or contrast shifts, partial occlusion, and sensor noise rather than only from idealized held-out examples. For compact models intended for deployment close to acquisition or analysis pipelines, clean performance alone is therefore insufficient if performance degrades sharply under such perturbations. In low-data clinical settings, demonstrating stability under such conditions is also important for the broader goal of trustworthy deployment.

This problem is especially relevant in light of recent efforts to benchmark robustness under medically meaningful corruptions. MedMNIST-C was introduced precisely to assess robustness under realistic image degradations---including Gaussian noise, blur, contrast adjustment, and compression artifacts---across MedMNIST-style biomedical benchmarks, reflecting the broader recognition that clean-test evaluation alone is not enough to characterize deployment behavior \cite{disalvo2024medmnistc}. In parallel, surveys of medical AI robustness continue to emphasize that adversarial fragility and robustness under distribution shift address different, but complementary, failure modes \cite{apostolidis2021survey, balendran2025robustness}. These perspectives motivate a direct robustness analysis of \method{} rather than an additional clean-only benchmark.

In this workshop paper, we address that question by treating robustness as the first focused extension of the \method{} research line. We retain the controlled MedMNIST setting used in the foundational manuscript and compare four scratch-trained compact backbones---ABMIL, Minimal-ViT, TransMIL, and \method{}---across seven datasets under four evaluation regimes: clean test data, common-corruption means, FGSM means, and PGD means. This design separates three practically relevant aspects of model behavior: baseline predictive performance, stability under realistic non-adversarial degradation, and resilience under gradient-based adversarial stress.

The resulting picture is consistent, although not uniform across settings. \method{} performs best overall on clean data and under common corruptions, which is the part of the robustness story most directly connected to likely deployment conditions. Under FGSM and PGD, performance declines sharply for all models, and no architecture is consistently best across all datasets. Even so, \method{} remains competitive and is the best-ranked transformer-based baseline on average under PGD. Taken together, the results do not overturn the original \method{} narrative. Rather, they refine it by showing that the architectural choices behind \method{} are not only associated with regime-dependent clean performance, but can also remain advantageous under realistic perturbation stress, whereas adversarial robustness remains a broader open problem.

\paragraph{Contributions.}
This paper makes three contributions. First, it presents the first robustness-focused extension of the recently introduced \method{} architecture by revisiting the MedMNIST benchmark under clean, corruption, FGSM, and PGD evaluation regimes. Second, it shows that \method{} consistently achieves strong mean-rank performance on clean data and under common corruptions across seven MedMNIST datasets, supporting a strong clean-corruption trade-off in a deployment-relevant setting. Third, it provides a cautious adversarial analysis: \method{} remains competitive under FGSM and PGD, but the results do not support broad adversarial-robustness claims, thereby identifying a clear direction for future work within the broader \method{} program.

\section{Related Work}
\paragraph{ZACH-ViT as a compact permutation-invariant backbone.}
The formal \method{} manuscript introduced the architecture as a compact Vision Transformer that removes both positional embeddings and the dedicated \texttt{[CLS]} token, replacing token-based aggregation with global average pooling over patch representations. The central claim of that work was not universal superiority, but \emph{regime-dependent inductive-bias alignment}: when spatial layout is weakly informative or inconsistent, reducing reliance on fixed spatial priors may be beneficial, whereas stronger positional support can become mildly useful as spatial structure increases. The present paper does not introduce a new architecture; instead, it examines how the same design behaves under robustness stress.

\paragraph{Compact transformers and MIL-style baselines in medical imaging.}
Transformer-based models in medical imaging include both compact end-to-end backbones and multiple-instance learning (MIL) formulations. TransMIL and ABMIL are especially relevant comparators because they also aggregate information across local representations, although they differ from \method{} in objective, architectural role, and intended use \cite{shao2021transmil, ilse2018abmil}. ABMIL employs attention-based pooling over instance features, while TransMIL models correlations between instances through a transformer architecture. These distinct aggregation philosophies provide substantive baselines for assessing whether any robustness advantage is specific to a compact permutation-invariant transformer or also emerges in MIL-based alternatives.

\paragraph{Robustness under realistic corruptions and adversarial attacks.}
Robustness in medical imaging is commonly studied along two complementary axes. Common corruptions approximate realistic variation arising from acquisition differences, compression, nuisance artifacts, and signal degradation \cite{hendrycks2019benchmark, disalvo2024medmnistc}, whereas adversarial attacks probe sensitivity to worst-case local perturbations \cite{goodfellow2015explaining, madry2018towards}. These settings are informative for different reasons and should not be treated as interchangeable. A model may be comparatively stable under realistic corruptions without being strongest under gradient-based attacks. In medical imaging specifically, adversarial fragility and broader robustness failures have both been recognized as meaningful barriers to reliable deployment \cite{apostolidis2021survey, balendran2025robustness}. Since compact medical imaging models are more likely to encounter acquisition variability than adaptive white-box attackers in routine use, both perspectives are informative, but they address different deployment questions.

\section{Methods}
\subsection{Backbone models}
We evaluate four scratch-trained compact baselines: ABMIL, Minimal-ViT, TransMIL, and \method{}. \method{} is the primary method of interest. As introduced in the foundational manuscript, it is a compact permutation-invariant transformer without positional embeddings or a dedicated class token, and it uses adaptive residual projections across dimensional transitions together with global average pooling over patch tokens. The comparison set was selected to remain consistent with the compact-model regime of the foundational \method{} study: all four models are trained from scratch and have parameter counts below 1M. This restriction keeps the comparison focused on low-capacity backbones that are relevant to low-data and deployment-oriented settings, while still spanning different design philosophies, including MIL-style aggregation and a more conventional compact transformer baseline.

\subsection{Datasets and few-shot protocol}
We retain the seven MedMNIST datasets used in the main \method{} study: BloodMNIST, PathMNIST, BreastMNIST, PneumoniaMNIST, DermaMNIST, OCTMNIST, and OrganAMNIST. Binary tasks are evaluated with AUC@0.5 and multi-class tasks with Macro-F1. All models are trained under the same few-shot protocol: 50 training samples per class, unchanged validation and test splits, batch size 16, 23 epochs, and five random seeds $\{3,5,7,11,13\}$. This low-data setup is intended to emphasize robustness under data scarcity rather than maximize performance through extensive tuning.

\subsection{Robustness evaluation}
Each model is evaluated under four conditions: (i) \textbf{Clean}, corresponding to performance on the original test set; (ii) \textbf{Corruption mean}, defined as the average performance across the evaluated common corruptions, namely Gaussian noise, Gaussian blur, brightness-contrast adjustment, JPEG compression, and cutout, each at three severity levels; (iii) \textbf{FGSM mean}, defined as the average performance across the evaluated FGSM perturbation strengths ($\epsilon \in \{1/255, 2/255, 4/255, 8/255\}$); and (iv) \textbf{PGD mean}, defined as the average performance across the evaluated PGD perturbation strengths using the same $\epsilon$ values, 10 attack steps, and step size $\epsilon/4$. Adversarial examples are clipped to the valid input range, and PGD additionally projects perturbed inputs back into the corresponding $L_\infty$ ball around the clean sample. For each dataset and condition, we report the mean and standard deviation over five seeds. These seed-level summaries provide empirical variability estimates, but they should not be interpreted as formal uncertainty quantification. To obtain dataset-agnostic summaries, we compute mean ranks across the seven datasets, where lower rank indicates better performance.

\noindent\textbf{Statistical considerations.}
To provide a basic indication of variability, we report mean and standard deviation across five random seeds. While the present study is not designed as a formal statistical comparison, the consistency of ranking patterns across datasets suggests that the observed differences are not driven by a single split. A more comprehensive statistical testing framework is left for future work.

\section{Results}
\subsection{The robustness extension preserves the original clean performance profile}
Table~\ref{tab:clean_corruption} shows that the clean performance profile remains broadly consistent with the original \method{} manuscript. The model performs particularly well on DermaMNIST and OCTMNIST, remains competitive on BloodMNIST and BreastMNIST, and stays close to the top-performing model on OrganAMNIST and PneumoniaMNIST. TransMIL remains strongest on PathMNIST, while ABMIL performs notably well on BreastMNIST. The most informative dataset-agnostic summary is the mean rank: \method{} attains the best overall clean rank (1.57), narrowly ahead of TransMIL (1.71), with ABMIL and Minimal-ViT trailing. This indicates that the robustness extension begins from the same central observation as the foundational paper---namely, that architectural alignment matters more than universal dominance across all regimes.

\subsection{ZACH-ViT shows the strongest overall profile under realistic corruptions}
\begin{figure*}[t]
    \centering
    \includegraphics[width=0.96\textwidth]{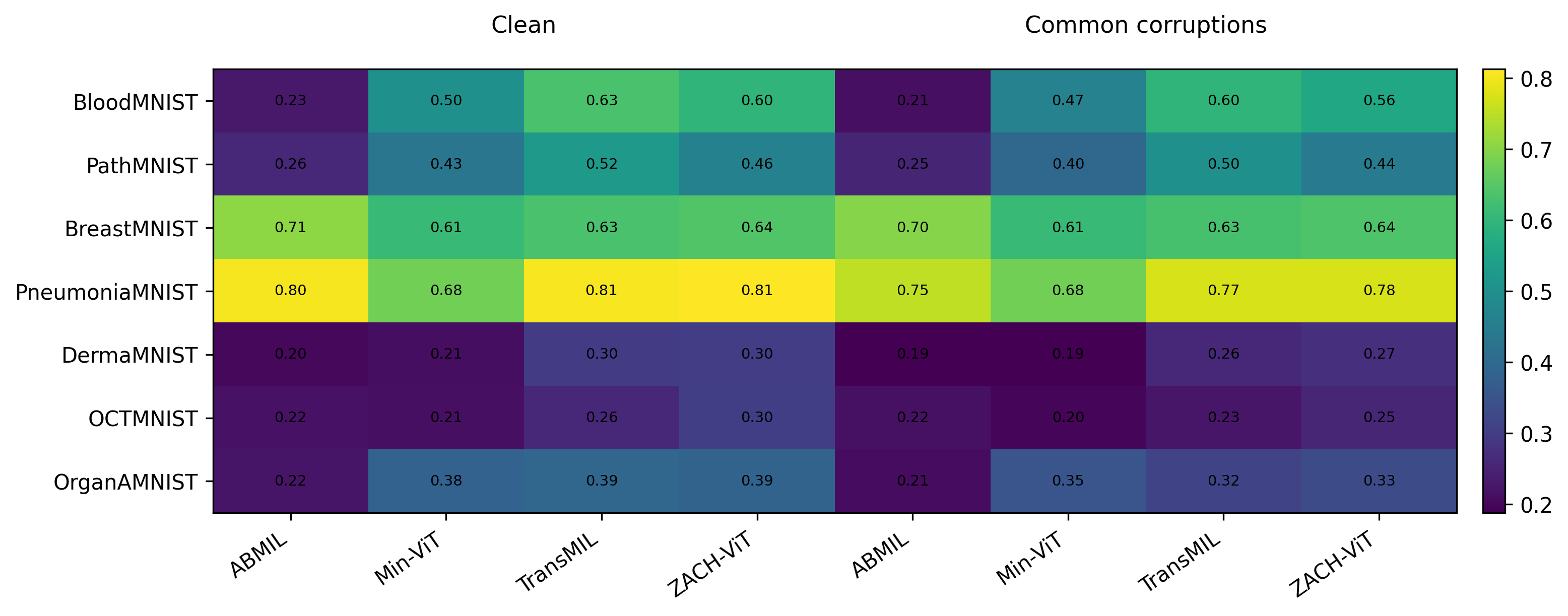}
    \caption{Clean performance and corruption-averaged robustness across the seven MedMNIST datasets. The clean performance pattern reported in the foundational \method{} study is largely preserved, and the corruption analysis further shows that \method{} provides the most favorable overall balance between baseline predictive performance and robustness to realistic image degradation.}
    \label{fig:clean_corr}
\end{figure*}

The main additional result of this study emerges under common corruptions. Figure~\ref{fig:clean_corr} and Table~\ref{tab:clean_corruption} show that \method{} again attains the best mean rank across the seven datasets (1.57), ahead of TransMIL (2.00), ABMIL (3.14), and Minimal-ViT (3.29). This pattern is not attributable to a single dataset. \method{} remains particularly strong on DermaMNIST, OCTMNIST, and PneumoniaMNIST, while staying close to the best model on BreastMNIST and OrganAMNIST. TransMIL retains a clear advantage on PathMNIST, but the broader pattern is consistent: the inductive-bias choice underlying \method{} remains favorable when robustness is defined in a deployment-relevant manner.

This result is the most practically relevant finding in the present study. In medical imaging, realistic failures are more likely to arise from compression, scanner variation, contrast changes, blur, noise, or partial occlusion than from carefully optimized white-box attacks. From that perspective, corruption robustness is not merely an auxiliary benchmark, but an important stress test for compact models intended for deployment-oriented evaluation. In our benchmark, this advantage is observed across multiple corruption families, including noise, blur, brightness-contrast shifts, JPEG compression, and cutout. The present results therefore strengthen the original \method{} claim by showing that the architectural advantages observed under clean evaluation can persist under plausible perturbations encountered in practice.

\subsection{Adversarial robustness remains limited for all models}
\begin{figure*}[t]
    \centering
    \includegraphics[width=0.96\textwidth]{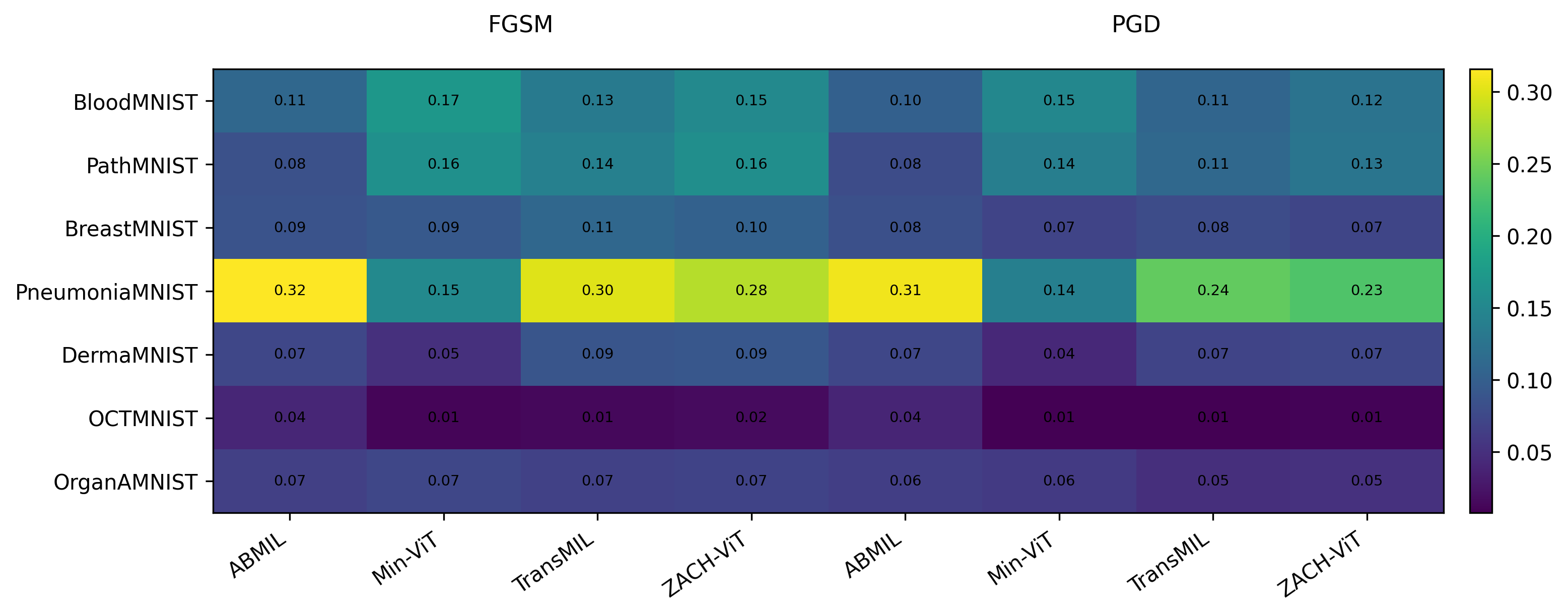}
    \caption{Adversarial stress testing under FGSM and PGD. All models degrade substantially relative to the clean setting. \method{} remains competitive and is the best-ranked transformer-based baseline on average under PGD, but no architecture is consistently best across all datasets.}
    \label{fig:attacks}
\end{figure*}

The adversarial results are more heterogeneous and should be interpreted carefully. Figure~\ref{fig:attacks} and Table~\ref{tab:attacks} show that performance declines sharply for every model under both FGSM and PGD. Under FGSM, \method{} attains the best overall mean rank (2.00), ahead of TransMIL (2.43), Minimal-ViT (2.57), and ABMIL (3.00). At the level of individual datasets, however, the strongest model varies: ABMIL performs best on BloodMNIST and PneumoniaMNIST, TransMIL on BreastMNIST, and \method{} on DermaMNIST and OCTMNIST. Under PGD, ABMIL is strongest overall (mean rank 2.00), whereas \method{} ranks second (2.29), remaining the best-ranked transformer-based baseline on average.

The most appropriate interpretation is therefore comparative rather than absolute. The present robustness extension does not support claims that \method{} is universally robust to adversarial attacks. Instead, it shows that the model remains among the stronger compact backbones when the threat model becomes substantially harsher. This interpretation is more restrained and more consistent with the empirical evidence. The architectural choices that appear beneficial for clean data and realistic corruptions do not, on their own, resolve white-box robustness.

\subsection{Corruption-specific severity plots support the multi-family robustness pattern}
\begin{figure*}[t]
    \centering
    \includegraphics[width=0.94\textwidth]{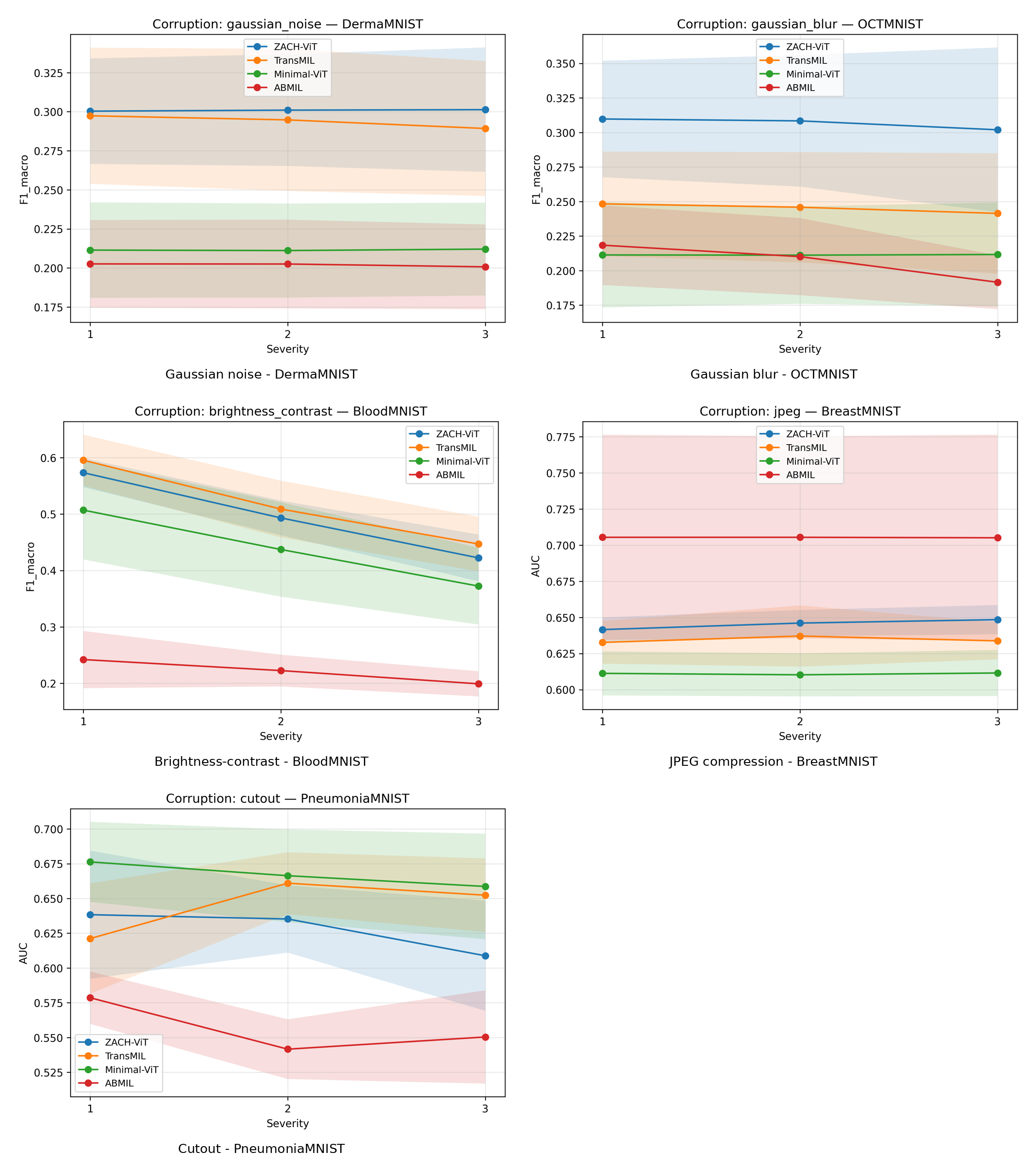}
    \caption{Representative corruption-specific severity plots from the benchmark outputs. The selected panels illustrate that the favorable corruption profile of \method{} is visible across multiple corruption families, including Gaussian noise, Gaussian blur, brightness-contrast adjustment, JPEG compression, and cutout.}
    \label{fig:corr_examples}
\end{figure*}

To complement the dataset-agnostic summaries, Figure~\ref{fig:corr_examples} shows representative corruption-specific severity plots drawn from the benchmark outputs. These examples provide direct evidence that the robustness pattern is not limited to a single degradation type or dataset. Across the selected examples, \method{} remains competitive as severity increases under Gaussian noise, Gaussian blur, brightness-contrast adjustment, JPEG compression, and cutout. The purpose of this figure is not to claim uniform dominance at every severity level, but to show that the favorable corruption profile of \method{} is visible across multiple corruption families rather than arising from a single artifact of averaging.

Mean rank is used as a dataset-agnostic summary to avoid over-interpreting absolute metric differences across heterogeneous tasks. We therefore interpret mean rank as a coarse comparative signal rather than a definitive measure of superiority.

\subsection{Mean-rank and retention summaries clarify the trade-off}
\begin{figure*}[t]
    \centering
    \includegraphics[width=0.92\textwidth]{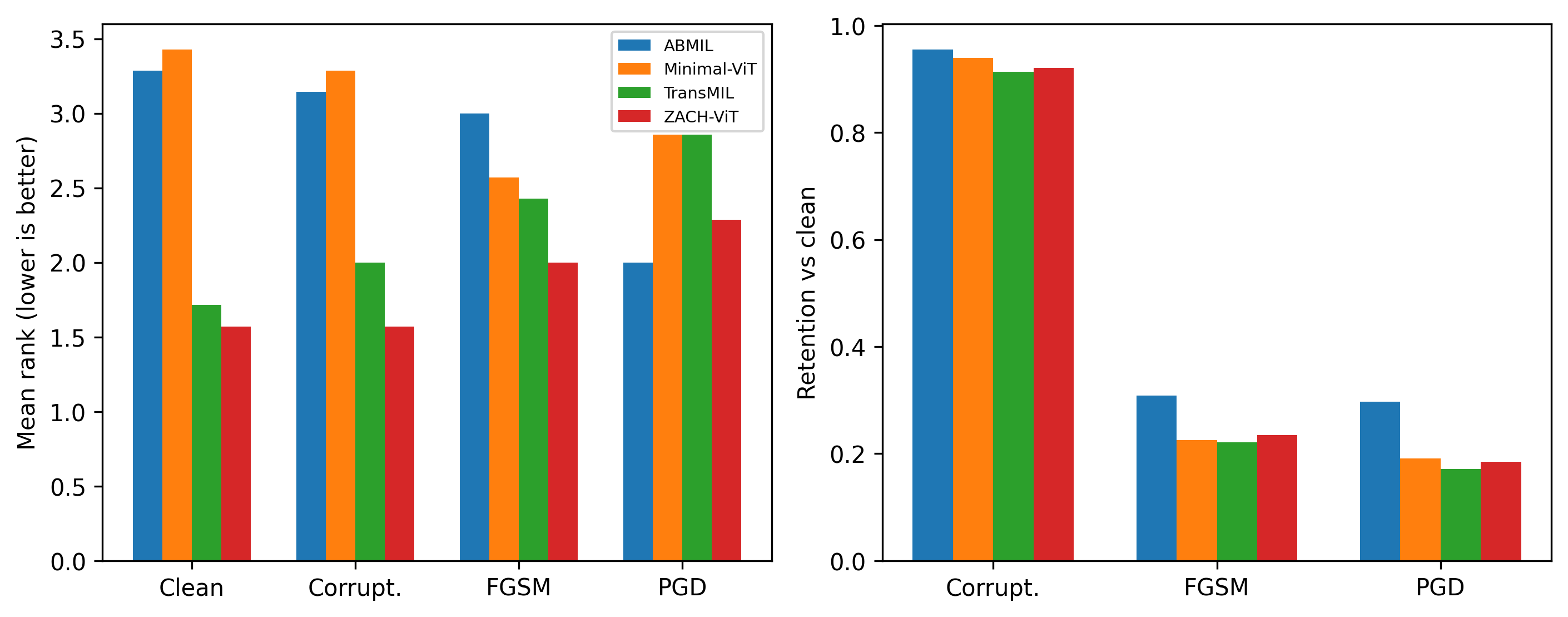}
    \caption{Dataset-agnostic summary of robustness behavior. \textbf{Left:} mean rank across clean, corruption, FGSM, and PGD regimes (lower is better). \textbf{Right:} average retention relative to each model's own clean performance. \method{} is the most balanced model overall, while ABMIL shows the strongest proportional retention under adversarial stress.}
    \label{fig:tradeoff}
\end{figure*}

Figure~\ref{fig:tradeoff} summarizes the benchmark from two complementary perspectives. The left panel shows mean ranks across the four evaluation regimes. \method{} ranks first on clean data and common corruptions, first under FGSM, and second under PGD. No other model matches that overall profile. The right panel reports retention relative to each model's own clean performance. Here ABMIL shows the highest proportional retention under common corruptions (0.96), FGSM (0.31), and PGD (0.30), but it also starts from a weaker clean baseline on most datasets. \method{} retains 0.92 of its clean performance under common corruptions, 0.23 under FGSM, and 0.18 under PGD, remaining close to or above the two transformer-based comparators.

This comparison clarifies the central message of the paper: robustness is not fully captured by a single scalar summary. A model may show high retention because it begins from a lower clean baseline, whereas another may still be preferable in absolute terms because its starting point is stronger. In the present benchmark, \method{} provides the best overall clean-corruption trade-off, while ABMIL deserves explicit credit for stronger adversarial retention.

\begin{table*}[t]
\centering
\caption{Clean and corruption-averaged performance across the seven MedMNIST datasets. The reported metric is AUC for binary tasks and Macro-F1 for multiclass tasks. Values are mean $\pm$ standard deviation over five seeds after averaging across corruption settings within each seed. We note that absolute differences between top-performing models are often small and should be interpreted with caution in the absence of formal hypothesis testing.}
\label{tab:clean_corruption}
\resizebox{\textwidth}{!}{%
\begin{tabular}{lcccccccc}
\toprule
& \multicolumn{4}{c}{Clean} & \multicolumn{4}{c}{Common corruptions} \\
\cmidrule(lr){2-5}\cmidrule(lr){6-9}
Dataset & ABMIL & Minimal-ViT & TransMIL & ZACH-ViT & ABMIL & Minimal-ViT & TransMIL & ZACH-ViT \\
\midrule
BloodMNIST & 0.232 $\pm$ 0.053 & 0.502 $\pm$ 0.079 & 0.634 $\pm$ 0.042 & 0.597 $\pm$ 0.025 & 0.214 $\pm$ 0.033 & 0.465 $\pm$ 0.073 & 0.596 $\pm$ 0.040 & 0.559 $\pm$ 0.023 \\
PathMNIST & 0.257 $\pm$ 0.028 & 0.432 $\pm$ 0.066 & 0.523 $\pm$ 0.067 & 0.461 $\pm$ 0.054 & 0.253 $\pm$ 0.031 & 0.398 $\pm$ 0.049 & 0.501 $\pm$ 0.097 & 0.443 $\pm$ 0.054 \\
BreastMNIST & 0.706 $\pm$ 0.071 & 0.611 $\pm$ 0.015 & 0.633 $\pm$ 0.016 & 0.642 $\pm$ 0.010 & 0.699 $\pm$ 0.064 & 0.611 $\pm$ 0.014 & 0.630 $\pm$ 0.011 & 0.639 $\pm$ 0.006 \\
PneumoniaMNIST & 0.804 $\pm$ 0.019 & 0.681 $\pm$ 0.031 & 0.808 $\pm$ 0.014 & 0.813 $\pm$ 0.012 & 0.753 $\pm$ 0.019 & 0.681 $\pm$ 0.029 & 0.774 $\pm$ 0.012 & 0.775 $\pm$ 0.010 \\
DermaMNIST & 0.202 $\pm$ 0.028 & 0.211 $\pm$ 0.029 & 0.298 $\pm$ 0.044 & 0.301 $\pm$ 0.035 & 0.188 $\pm$ 0.023 & 0.188 $\pm$ 0.025 & 0.259 $\pm$ 0.042 & 0.272 $\pm$ 0.035 \\
OCTMNIST & 0.219 $\pm$ 0.024 & 0.213 $\pm$ 0.039 & 0.258 $\pm$ 0.024 & 0.304 $\pm$ 0.041 & 0.217 $\pm$ 0.027 & 0.196 $\pm$ 0.028 & 0.226 $\pm$ 0.030 & 0.255 $\pm$ 0.028 \\
OrganAMNIST & 0.224 $\pm$ 0.025 & 0.384 $\pm$ 0.041 & 0.394 $\pm$ 0.024 & 0.388 $\pm$ 0.027 & 0.209 $\pm$ 0.020 & 0.353 $\pm$ 0.024 & 0.316 $\pm$ 0.025 & 0.334 $\pm$ 0.016 \\
\midrule
Mean rank & 3.29 & 3.43 & 1.71 & 1.57 & 3.14 & 3.29 & 2.00 & 1.57 \\
\bottomrule
\end{tabular}%
}
\end{table*}
\begin{table*}[t]
\centering
\caption{FGSM- and PGD-averaged adversarial performance across the seven MedMNIST datasets. Values are mean $\pm$ standard deviation over five seeds after averaging across attack strengths within each seed. FGSM uses $\epsilon \in \{1/255, 2/255, 4/255, 8/255\}$. PGD uses the same $\epsilon$ values, 10 steps, and step size $\epsilon/4$. We note that absolute differences between top-performing models are often small and should be interpreted with caution in the absence of formal hypothesis testing.}
\label{tab:attacks}
\resizebox{\textwidth}{!}{%
\begin{tabular}{lcccccccc}
\toprule
& \multicolumn{4}{c}{FGSM} & \multicolumn{4}{c}{PGD} \\
\cmidrule(lr){2-5}\cmidrule(lr){6-9}
Dataset & ABMIL & Minimal-ViT & TransMIL & ZACH-ViT & ABMIL & Minimal-ViT & TransMIL & ZACH-ViT \\
\midrule
BloodMNIST & 0.110 $\pm$ 0.005 & 0.170 $\pm$ 0.015 & 0.134 $\pm$ 0.021 & 0.152 $\pm$ 0.018 & 0.102 $\pm$ 0.006 & 0.150 $\pm$ 0.015 & 0.108 $\pm$ 0.014 & 0.125 $\pm$ 0.020 \\
PathMNIST & 0.084 $\pm$ 0.010 & 0.162 $\pm$ 0.010 & 0.142 $\pm$ 0.026 & 0.160 $\pm$ 0.037 & 0.079 $\pm$ 0.008 & 0.139 $\pm$ 0.011 & 0.111 $\pm$ 0.021 & 0.127 $\pm$ 0.036 \\
BreastMNIST & 0.086 $\pm$ 0.064 & 0.094 $\pm$ 0.021 & 0.111 $\pm$ 0.056 & 0.103 $\pm$ 0.038 & 0.082 $\pm$ 0.061 & 0.071 $\pm$ 0.019 & 0.079 $\pm$ 0.050 & 0.071 $\pm$ 0.031 \\
PneumoniaMNIST & 0.316 $\pm$ 0.070 & 0.153 $\pm$ 0.055 & 0.301 $\pm$ 0.061 & 0.282 $\pm$ 0.015 & 0.309 $\pm$ 0.068 & 0.141 $\pm$ 0.051 & 0.242 $\pm$ 0.047 & 0.231 $\pm$ 0.011 \\
DermaMNIST & 0.074 $\pm$ 0.026 & 0.050 $\pm$ 0.021 & 0.089 $\pm$ 0.011 & 0.091 $\pm$ 0.015 & 0.073 $\pm$ 0.026 & 0.043 $\pm$ 0.018 & 0.070 $\pm$ 0.011 & 0.074 $\pm$ 0.015 \\
OCTMNIST & 0.041 $\pm$ 0.010 & 0.012 $\pm$ 0.007 & 0.014 $\pm$ 0.006 & 0.017 $\pm$ 0.011 & 0.040 $\pm$ 0.010 & 0.008 $\pm$ 0.006 & 0.008 $\pm$ 0.005 & 0.011 $\pm$ 0.007 \\
OrganAMNIST & 0.065 $\pm$ 0.020 & 0.073 $\pm$ 0.022 & 0.067 $\pm$ 0.016 & 0.070 $\pm$ 0.013 & 0.065 $\pm$ 0.020 & 0.060 $\pm$ 0.019 & 0.049 $\pm$ 0.016 & 0.051 $\pm$ 0.007 \\
\midrule
Mean rank & 3.00 & 2.57 & 2.43 & 2.00 & 2.00 & 2.86 & 2.86 & 2.29 \\
\bottomrule
\end{tabular}%
}
\end{table*}

\section{Discussion}
The main contribution of this paper is that it extends the formal introduction of \method{} in a way that remains consistent with the original scientific claim. The foundational manuscript argued that architectural performance should be interpreted through inductive-bias alignment with data structure, rather than through universal benchmark dominance alone. Within that framework, removing fixed positional priors can be beneficial when spatial structure is weakly informative or unstable. The present results suggest that this advantage is not limited to clean held-out evaluation. It also carries over, to a meaningful extent, to robustness under realistic corruptions, precisely the regime in which compact deployment-oriented models are likely to encounter difficulties.
A controlled ablation isolating the effect of permutation invariance on robustness remains an important direction for future work.

This is relevant for the broader development of \method{} as a research line. A foundational architecture paper becomes more informative when its first extension is not simply another clean benchmark, but a focused stress test that clarifies where the design is and is not advantageous. In that sense, the current study refines the identity of \method{}: the model is not defined by dominance in every evaluation regime, but by a balanced combination of compactness, clean predictive performance, and corruption stability in low-data medical imaging. One plausible explanation is that permutation-invariant aggregation reduces reliance on brittle or acquisition-dependent positional correlations when medically relevant signal is distributed rather than tightly anchored to fixed spatial structure, encouraging greater reliance on local visual evidence and compositional statistics, although the present study was not designed to isolate this mechanism directly.

The corruption results also suggest a plausible connection to Edge AI and deployment-oriented medical imaging. Edge and near-device inference are often motivated by latency, connectivity, and privacy constraints in healthcare settings \cite{rocha2024edgeai, lakshminarayanan2023edge}. Within that broader context, a compact model that performs well under realistic corruptions is a reasonable candidate backbone for further evaluation in production-oriented and edge-aware clinical AI pipelines. Our results therefore support \method{} as a compact backbone that may be relevant to edge-oriented medical imaging workflows, especially when robustness to acquisition variability is important. This interpretation should, however, remain bounded: the present study does not evaluate inference latency, memory footprint, energy use, thermal constraints, integration into medical-device software, or prospective clinical performance. Those aspects would require dedicated hardware and translational validation before any deployment claim could be justified.

For the PHAROS community, where the intersection of healthcare AI and practical deployment is central, this study offers two takeaways: (1) architectural choices that appear beneficial in clean evaluation can also confer robustness to realistic acquisition variability, contributing to the development of more trustworthy AI systems; and (2) adversarial robustness remains a distinct challenge requiring dedicated methods beyond inductive-bias tuning. Although this study does not address fairness explicitly, robustness in low-data medical settings should ultimately be evaluated together with subgroup reliability and equitable performance under deployment shift.

At the same time, the adversarial results impose important constraints on interpretation. Neither \method{} nor any comparator is robust in an absolute sense under FGSM or PGD. A more appropriate conclusion is that the architecture remains competitive under attack while showing clearer advantages under realistic perturbations. That is already a meaningful result for a workshop paper.

\section{Limitations}
This study has several limitations. First, the benchmark is intentionally restricted to scratch-trained compact baselines and does not include large pretrained foundation models or transfer learning settings, which are commonly used in medical imaging. The present study therefore focuses on controlled low-data behavior rather than state-of-the-art performance under large-scale pretraining. Second, robustness is summarized through averages over corruption types and attack strengths, which is suitable for a comparative workshop study but does not isolate the source of each performance difference. Third, the adversarial analysis is empirical rather than certified and should be interpreted as stress testing rather than formal robustness assurance. In addition, the adversarial evaluation is limited to standard FGSM and short-horizon PGD attacks and should not be interpreted as a comprehensive robustness assessment under stronger or adaptive threat models. Fourth, all datasets are drawn from MedMNIST, which provides a useful controlled benchmark but does not substitute for external validation, prospective evaluation, or silent deployment testing in intended-use environments \cite{yang2023medmnist}. Fifth, we do not assess subgroup fairness or demographic performance disparities, in part because such annotations are not available in the present benchmark. Finally, the Edge AI relevance discussed above is inferential rather than demonstrated, because no hardware-level measurements or workflow integration experiments are included. 

\section{Conclusion}
We presented the first robustness-focused extension of the recently introduced \method{} architecture for low-data medical imaging. Across seven MedMNIST datasets and four evaluation regimes, \method{} achieved the best mean rank on clean data and under common corruptions while remaining competitive under FGSM and PGD attacks. The present study therefore extends, rather than replaces, the original \method{} narrative: design choices motivated by inductive-bias alignment with spatial structure can provide an attractive balance of predictive performance, realistic robustness, and efficiency in data-constrained medical imaging, even though adversarial robustness remains incomplete. For PHAROS and related healthcare-AI venues, these findings position \method{} as a principled compact backbone whose strengths are most visible under realistic perturbation stress, and they motivate further evaluation in edge-oriented medical imaging settings. The corruption/adversarial evaluation protocol presented here may also serve as a useful template for other PHAROS participants assessing compact medical imaging backbones.


{\small
\begin{thebibliography}{99}\setlength{\itemsep}{0pt}

\bibitem{angelakis2026zachvit}
Athanasios Angelakis.
\newblock ZACH-ViT: Regime-dependent inductive bias in compact vision transformers for medical imaging.
\newblock arXiv:2602.17929, 2026.

\bibitem{balendran2025robustness}
Alan Balendran, Céline Beji, Florie Bouvier, Ottavio Khalifa, Theodoros Evgeniou, Philippe Ravaud, and Raphaël Porcher.
\newblock A scoping review of robustness concepts for machine learning in healthcare.
\newblock \emph{npj Digital Medicine}, 8(1):38, 2025.

\bibitem{disalvo2024medmnistc}
Francesco Di Salvo, Sebastian Doerrich, and Christian Ledig.
\newblock MedMNIST-C: Comprehensive benchmark and improved classifier robustness by simulating realistic image corruptions.
\newblock arXiv:2406.17536, 2024.

\bibitem{apostolidis2021survey}
Kyriakos D. Apostolidis and George A. Papakostas.
\newblock A survey on adversarial deep learning robustness in medical image analysis.
\newblock \emph{Electronics}, 10(17):2132, 2021.

\bibitem{shao2021transmil}
Zhuchen Shao, Hao Bian, Yang Chen, Yifeng Wang, Jian Zhang, Xiangyang Ji, and Yongbing Zhang.
\newblock TransMIL: Transformer based correlated multiple instance learning for whole slide image classification.
\newblock In \emph{Advances in Neural Information Processing Systems}, 2021.

\bibitem{ilse2018abmil}
Maximilian Ilse, Jakub M. Tomczak, and Max Welling.
\newblock Attention-based deep multiple instance learning.
\newblock In \emph{International Conference on Machine Learning}, 2018.

\bibitem{hendrycks2019benchmark}
Dan Hendrycks and Thomas Dietterich.
\newblock Benchmarking neural network robustness to common corruptions and perturbations.
\newblock In \emph{International Conference on Learning Representations}, 2019.

\bibitem{goodfellow2015explaining}
Ian J. Goodfellow, Jonathon Shlens, and Christian Szegedy.
\newblock Explaining and harnessing adversarial examples.
\newblock In \emph{International Conference on Learning Representations}, 2015.

\bibitem{madry2018towards}
Aleksander Madry, Aleksandar Makelov, Ludwig Schmidt, Dimitris Tsipras, and Adrian Vladu.
\newblock Towards deep learning models resistant to adversarial attacks.
\newblock In \emph{International Conference on Learning Representations}, 2018.

\bibitem{yang2023medmnist}
Jiancheng Yang, Rui Shi Huang, Jiajing Li, et al.
\newblock MedMNIST v2: A large-scale lightweight benchmark for 2D and 3D biomedical image classification.
\newblock \emph{Scientific Data}, 10:48, 2023.

\bibitem{rocha2024edgeai}
Atslands Rocha, Matheus Monteiro, César Mattos, Madson Dias, Jorge Soares, Regis Magalhães, and José Macedo.
\newblock Edge AI for Internet of Medical Things: A literature review.
\newblock \emph{Computers \& Electrical Engineering}, 116:109202, 2024.

\bibitem{lakshminarayanan2023edge}
Vishal Lakshminarayanan, Aswathy Ravikumar, Harini Sriraman, Sujatha Alla, and Vijay Kumar Chattu.
\newblock Health Care Equity Through Intelligent Edge Computing and Augmented Reality/Virtual Reality: A systematic review.
\newblock \emph{Journal of Multidisciplinary Healthcare}, 16:2839--2859, 2023.

\end{thebibliography}}
\end{document}